\documentclass[12pt,a4paper]{article}

\usepackage{float}
\usepackage{url}

\usepackage[title]{appendix}
\usepackage{hyperref}
\usepackage{amsmath}
\usepackage{amssymb}
\usepackage{mathtools}
\usepackage{graphicx}
\usepackage{threeparttable}
\usepackage{adjustbox,lipsum}
\usepackage{algorithm}
\usepackage[noend]{algpseudocode}

\usepackage[a4paper,left=2.5cm,right=2.5cm,top=2.5cm,bottom=2.5cm]{geometry}

\begin{document}
\title{Revisiting randomized choices in isolation forests}
\author{David Cortes}
\maketitle

\begin{abstract}
Isolation forest or "iForest" is an intuitive and widely used algorithm for anomaly detection that follows a simple yet effective idea: in a given data distribution, if a threshold (split point) is selected uniformly at random within the range of some variable and data points are divided according to whether they are greater or smaller than this threshold, outlier points are more likely to end up alone or in the smaller partition. The original procedure suggested the choice of variable to split and split point within a variable to be done uniformly at random at each step, but this paper shows that "clustered" diverse outliers - oftentimes a more interesting class of outliers than others - can be more easily identified by applying a non-uniformly-random choice of variables and/or thresholds. Different split guiding criteria are compared and some are found to result in significantly better outlier discrimination for certain classes of outliers.
\end{abstract}

\section{Introduction}

The isolation forest (a.k.a. \textsc{iForest}, \cite{iso}) algorithm for anomaly detection tries to exploit the idea that outliers, by definition, are "few and different" from normal data points. The premise behind the algorithm is very intuitive: in a given distribution of data points $\mathbf{X} \in \mathbb{R}^{m \times n}$, if picking a variable (column) at random within the data, if one selects a value uniformly at random within the range of the data distribution for that variable and then groups observations (rows) according to whether they are greater or smaller than this threshold across the chosen variable, then outliers (anomalies) have a larger chance of ending up alone or in a smaller group when compared to normal observations.

The algorithm exploits this idea by performing recursive partitions of the data  - that is, a given set or subset of points is first split according to this random choice, then at each branch or group that this division produces, the procedure is repeated again but this time considering only the points that reached this subset, and it continues doing so until each point becomes isolated (is the only point present in a given division) or until some other termination criteria is met. Under this scheme, outlier points will in expectation take fewer steps/divisions/branchings to become isolated, and thus the expected isolation depth can be used as an anomaly score for each point (the shorter, the more anomalous).

For better results, this procedure is repeated many times (suggestion in \cite{iso} was 100 times), with each different run (from here on, a "decision tree" or simply "tree") taking only a small sub-sample of the data (suggestion in \cite{iso} was to sub-sample 256 points without replacement for each tree), with the observed isolation depth for each observation averaged across trees  - the set of which form a "forest" - and compared against the expected isolation depth if all the data were distributed uniformly at random across all dimensions.

For a more computationally efficient implementation, the procedure may be stopped before reaching isolation of every point at a given tree, since outliers can only be considered to be so if their isolation depth is below average, and thus a remainder can be extrapolated for points that reach a non-isolated node with depth greater or equal than the expected average for uniformly-random data. The expected isolation depth (and thus the extrapolated remainder) can be approximated efficiently through a closed-form formula (see \cite{depth}), resulting in a very fast procedure compared to e.g. distance-based algorithms such as \textsc{LOF} ("local outlier factor", \cite{lof}).

The isolation forest algorithm as-is has enjoyed wide success and widespread usage, achieving good results in a variety of benchmarks encompassing different types of outliers (see e.g. \cite{genif}) and usually outperforming other model classes, although some issues have been identified with the logic behind the algorithm (e.g. in \cite{inne}). Many subsequent works have built upon the idea to create better variations of the procedure, some of which will be analyzed in the next sections.

While overall useful as a general outlier detector and perhaps one of the best all-around methods as measured by aggregated metrics across benchmarks, a deeper look at performance comparisons in the literature would reveal that \textsc{iForest} and modifications thereof typically outperform other classes of methods in certain types of datasets, but not in all of them. This work tries to characterize these cases and proposes a different guiding heuristic for split selection in \textsc{iForest}, which is compared against similar ideas and found to outperform them in datasets in which isolation-based methods dominate others, at the expense of hindered performance in datasets in which non-tree-based methods outperform \textsc{iForest}.

\section{Related work}

There are many potential ways in which the isolation forest procedure can be improved upon while following the same core concept behind it. For example, \cite{ext} highlights biases that are introduced by the way that data splits are created and shows some simple ways in which the procedure can fail to produce adequate results, proposing a simple fix (the "extended isolation forest" algorithm or \textsc{EIF}) which, instead of producing splits by one variable at a time (or "axis-parallel splits"), produces split thresholds with respect to randomly-generated hyperplanes, which encompass only a couple of variables at a time.

The same extension is taken a step further in the "split criterion isolation forest" or \textsc{SCiForest} from \cite{sci}, proposing not only doing the splits by random hyperplanes, but also determining the split threshold by a deterministic criterion that aims at making each tree branch (groups that are obtained after dividing points according to the threshold) more homogeneous, and reporting increased performance metrics in typical anomaly detection datasets from these changes.

The "robust random cut forest" or \textsc{RRCF} from \cite{rrcf} proposed choosing the variable to split with a probability proportional to the range that it spans instead of choosing it uniformly at random, following the idea that variables with a wider ranger are more important in a multivariate distribution and more likely to differentiate outliers.

Other models such as the "density estimation trees" or \textsc{DET} from \cite{det} also propose more elaborate criteria for choosing the variable and the split point, in the case of \cite{det} considering the distribution of all variables rather than just a subset of them as in \cite{sci}, and choosing both the variable to split and the split point within the variable according to a deterministic criterion that aims at grouping together more points where the multivariate density of the data is higher, with density viewed as the amount of points per volume unit in the space spanned by the features. The "generalized isolation forest" (\textsc{GIF}) from \cite{genif} and the "one-class random forest" (\textsc{OCRF}) from \cite{def} also proposed similar ideas, but with slightly different splitting criteria.

\section{Different types of outliers}

So far, outliers or anomalies have only been characterized here as being "few and different" - for example, in typical datasets for outlier detection, outliers are sampled from a minority group which differs in some fundamental aspect from the majority, such as cross-sectional measurements of soil samples in which outliers are measurements taken from a different type of soil than the rest. This nevertheless does not provide the full picture, as "outliers" or "anomalies" in a dataset can be generated by different events or processes, such as incorrect data entry or attempted fraudulent purchases.

A lot of literature has been written about "extreme value" identification in univariate distributions, which is one type of anomaly in which the value for some particular variable in an observation is very different from the rest, and can be caused e.g. by incorrect data entry that types an additional digit. More interesting classes of anomalies however typically involve relationships between many variables, with outliers not necessarily having extreme values in any particular variable, but presenting relationships between them which do not match with the rest. Most of the "anomaly detection" or "one-class classification" literature as well as public datasets focus on these cases.

A distinction is made in \cite{sci} between "scattered" and "clustered" outliers, with the "clustered" outliers being considered as more "interesting", as they are classes of outliers which typically originate through some repeated process (such as fraudulent activity) as opposed to "scattered" outliers which originate from unrelated events. These "clustered" outliers are deemed harder to identify due to aspects such as "masking", which \cite{sci} tries to address in its proposed method.

Most of the literature nevertheless makes no distinctions between classes of outliers. Oftentimes, a new method is introduced, and its utility demonstrated by identifying a pattern of anomality in randomly-generated data and aggregating performance metrics across different datasets, evaluating the method as a general anomaly detector. Such a presentation however provides an incomplete depiction, as different methods make a trade-off in being better at identifying certain classes of outliers at the expense of being worse at identifying others as will be shown in the following sections, but such trade-off is rarely discussed.

For example, methods based on density or likelihood such as \textsc{DET} will determine outlierness according to how uncommon a combination of values would be, which in a multi-modal distribution would give a more anomalous value to observations that cluster near one of the non-majority modes compared to those that cluster near the largest mode, while methods based on local outlierness or neighborhoods such as \textsc{iNNE} from \cite{inne} would not deem observations near minority modes any less anomalous than those near majority modes. For outliers such as those in the "Satellite" dataset\footnote{\url{http://odds.cs.stonybrook.edu/satellite-dataset/}} for example, the latter might be a disadvantage, while for outliers in the "ForestCover" dataset\footnote{\url{http://odds.cs.stonybrook.edu/forestcovercovertype-dataset/}} such a property would be desirable.

This work shows that there is no "silver bullet" and suggests that different methods should be employed for different outlier types or desired anomalous patterns to catch.

While many previous works have tried to make a distinction between "global" and "local" outliers, there are other perhaps more important aspects which could be used to categorize them. In the public datasets that are typically used to compare algorithms (from e.g. \cite{odds} or \cite{repo2}, which have been used in e.g. \cite{iso}, \cite{sci}, \cite{genif}, \cite{ocrf}, and many others), one could also make a distinction according to the data generation process - for example:
\begin{enumerate}
\item Imbalanced binary classification, in which anomalies are all from a different class to which less than 1\% of the observations belong. These are in general the easiest datasets to simulate. Datasets such as as "Pima", "ForestCover" and "Mnist" belong to this group.
\item Rare and "scattered" outliers, typically from multi-class classification, in which very few points belonging to different minority classes are added to a majority class or classes. Datasets such as "ALOI" belong to this group.
\item Multi-label classification in which minority classes are grouped together as outliers, typically representing a much larger percentage of the total than in the imbalanced classification scenario, and the non-anomalous cases sometimes being a mixture of different majority classes. Datasets such as "Satellite", "Arrhythmia" and "Annthyroid" belong to this group.
\end{enumerate}

This is not by any means a comprehensive categorization (one could also make a distinction according to e.g. variable distributions), and not all datasets or outlier types belong to any of these groups - for instance, the "Pendigits" dataset is a multi-label classification scenario in which outliers are a single down-sampled class while inliers are the non-downsampled classes. Other datasets such as "SpamBase" could be considered a mixture of minority-classes and scattered outliers. Practical applications with non-artificial outliers are likely to encompass a mixture of different such scenarios, with areas like fraud detection being likely more similar to the "SpamBase" case than to the "ForestCover", for example.

When looked from this perspective, it becomes logical to suspect that e.g. highly-local methods might be very suitable for the second group, but not for the third group, while not-so-granular methods such as the linear models from \cite{ocsvm} could be very suitable for the first group. The first and third group could both be considered to be "clustered" outliers, and both are likely to be the cases of utmost interest in different types of applications, but methods that do well for one might not do so well for the other.

If taking \textsc{iForest} as a base, a quick experiment would show that different ways of choosing the split threshold or the variable to split would present visible trade-offs across datasets - among others, one could think of using kurtosis as a way to choose variables in order to catch extreme-valued outliers (in this experiment, by picking columns with a probability proportional to their kurtosis), or use an information gain criterion to determine the split threshold in the chosen variable similarly to \cite{imputer} (more on this method to follow in the next sections):
\begin{table}[H]
\centering
\caption {iForest with different feature and split selection strategies}
\begin{adjustbox}{max width=\textwidth}{\centering
\begin{tabular}{|r|c|c|c|}
 \hline
   &   \multicolumn{3}{|c|}{ \textbf{Area under the ROC curve}} \\
 \textbf{Guiding heuristic} & \textbf{Satellite} & \textbf{Annthyroid} & \textbf{Pendigits} \\
 \hline
Uniformly random & 0.718 & 0.827 & 0.957 \\  \hline
Kurtosis & 0.711 & 0.979 & 0.945 \\  \hline
Pooled gain & 0.857 & 0.829 & 0.948 \\  \hline
\end{tabular}}\end{adjustbox}
\end{table}

Both of which would lead to improved outlier detection in one dataset but not in the other two, thus making it impossible to say that one configuration is better than the others as a general outlier detector, but suggesting that a single universal outlier detector might not be the best approach either.

The overall hardest classes of outliers are perhaps those in which variable distributions are multi-modal and outliers are "clustered" around many different minority modes of different combinations of variables (such as the "Satellite" dataset). These are coincitendally the cases in which \textsc{iForest} tends to outperform other families of methods, and will be the focus of interest here, at the expense of other classes of outliers.

\section{Non-uniformly-random trees}

\textsc{DET} and \textsc{SCiForest} both suggest that carefully-chosen split points could result in improved outlier detectors compared to uniformly-random splits, but the metrics they aim at maximizing are quite different and lead to very dissimilar types of data partitions and tree structures.

The \textsc{SCiForest} model, looked from a slightly different perspective, will at each step first generate a random linear combination of a few variables (suggestion was two variables at a time), and then look for the split threshold that minimizes the average of the standard deviation of this linear combination of variables in  each side of the split. For a better-quality partition, \cite{sci} suggests trying many different linear combinations (suggestion is 10) at each step, and standardizing these numbers to make them comparable, proposing a "gain" criterion in a similar spirit as in supervised decision tree algorithms:
$$
\text{gain}_{\text{avg}} = \frac{\sigma_{\text{all}} - \frac{\sigma_{\text{left}} + \sigma_{\text{right}}}{2}}{\sigma_{\text{all}}}
$$
Where $\sigma_{\text{left}}$ and $\sigma_{\text{right}}$ are the standard deviations observed in the distribution of the randomly-generated linear combination among the points that go to each side of the comparison threshold ("less than" and "greater than") set upon this same linear combination, and $\sigma_{\text{all}}$ is the standard deviation among these points before partitioning them.

This split criterion will tend to favor partitions in which only one or a few points are put into a single branch while the majority go to the other, leading to quick isolation of extreme-valued points but perhaps generating insufficient discrimination among non-extreme values. As many points quickly start becoming isolated in earlier splits, it takes relatively fewer splits to reach the same termination criterion used by "iForest", resulting in smaller trees (containing fewer nodes) and thus in principle a computationally-efficient procedure if the termination criteria remains the same. It should be noted however that the expected isolation depth for uniformly-random data in this case would not be the same as for the simpler \textsc{iForest}, but would rather be a usually larger number, given by:
$$
\mathbb{E}[d(m)] = \frac{(\sum_{i=1}^{m} i) - 1}{m}
$$
Where $m$ is the number of uniformly-random points. This comes from the fact that the optimal split for uniformly-random data corresponds to leaving a single observation in one branch and the rest in the other.

\begin{figure}[H]
\centerline{\includegraphics[totalheight=2.7cm]{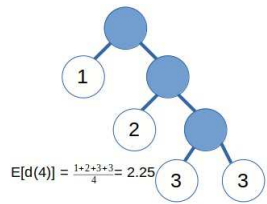}}
    \caption{Expected isolation depth for 4 uniform points under averaged gain}
    \label{fig:verticalcell1}
\end{figure}

A small experiment with the "Satellite" dataset reveals that the expected isolation depth from \textsc{iForest} might indeed not be as appropriate for \textsc{SCiForest}:
\begin{figure}[H]
\centerline{\includegraphics[totalheight=6cm]{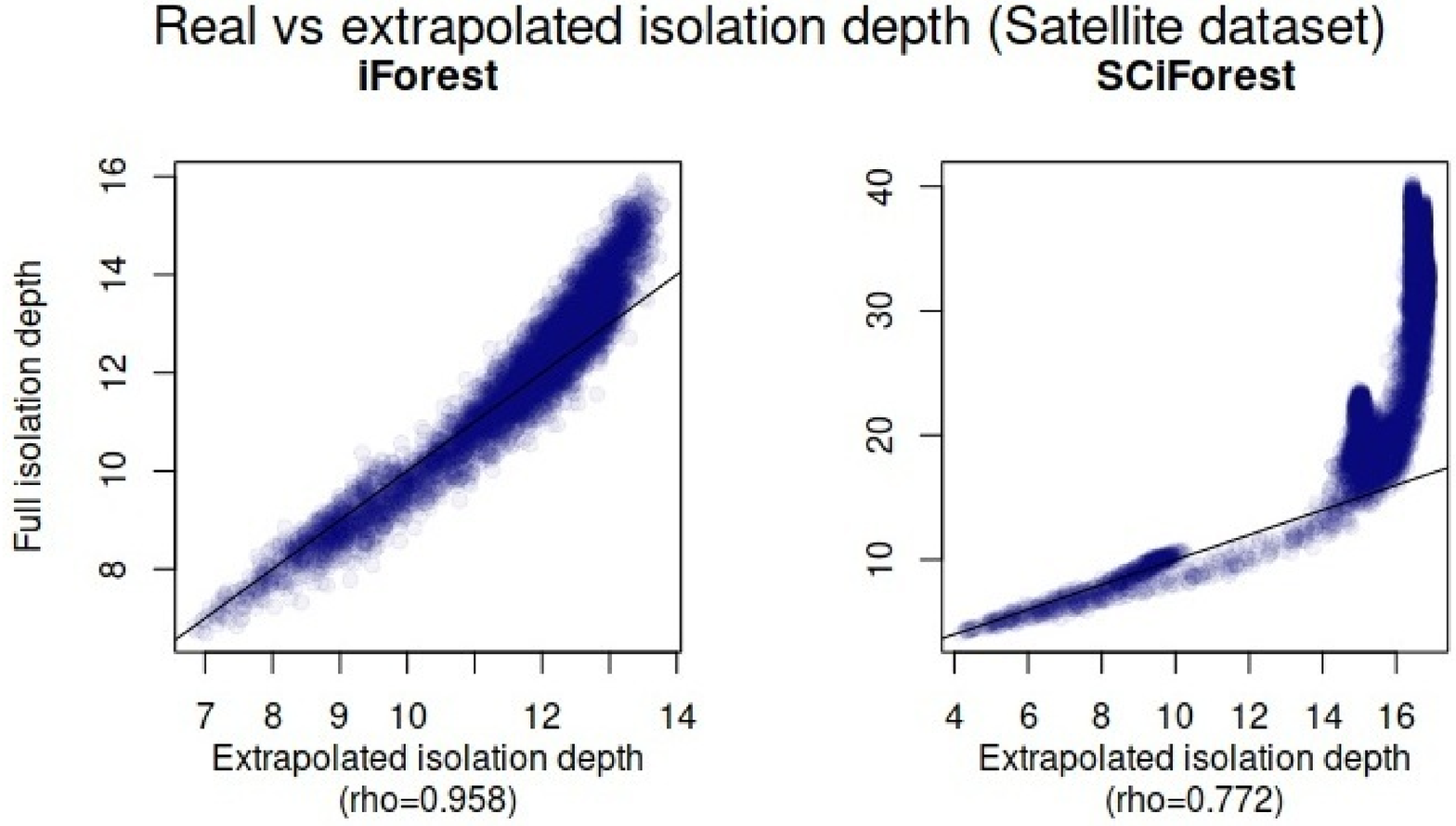}}
    \label{fig:verticalcell2}
\end{figure}

Nevertheless, \cite{sci} kept the same threshold and remainder extrapolation as in the original \textsc{iForest} in their design.

When data distributions are multi-modal and these modes very separable, this averaged gain criterion can tend to result in quick separation of groups, but at the expense of so-called "ghost regions" (see \cite{ext}), the moreso if a remaining depth is extrapolated instead of running the procedure until full isolation:
\begin{figure}[H]
\centerline{\includegraphics[totalheight=5.5cm]{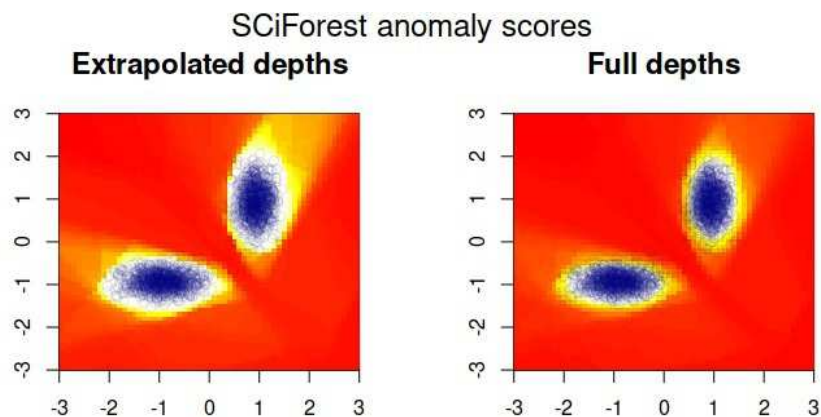}}
	\caption{Anomaly scores from \textsc{SCiForest} in randomly-generated bimodal data, with no sub-sampling to make the effects more visible. The redder, the higher the anomaly score.}
    \label{fig:verticalcell3}
\end{figure}

Perhaps counterintuitively, the splits generated following this criterion seem to be less sensitive to the presence of local outliers in the training data compared to uniformly-random choices:
\begin{figure}[H]
\centerline{\includegraphics[totalheight=7cm]{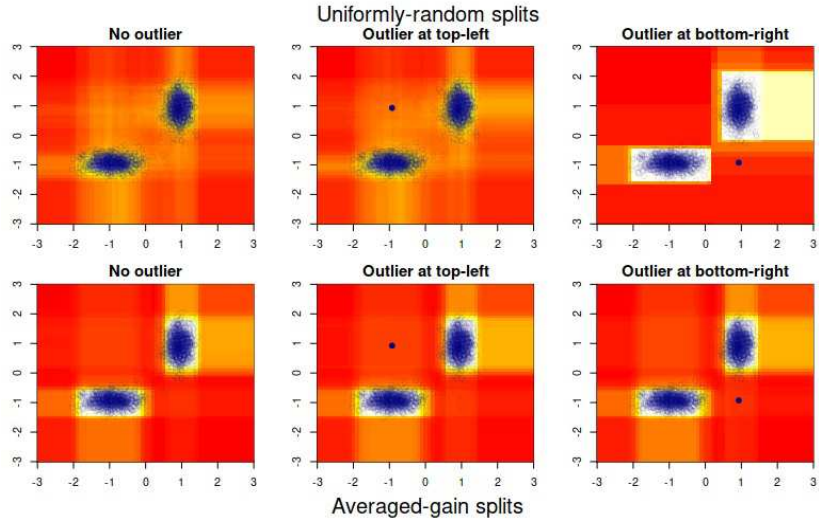}}
	\caption{Influence of an outlier in the splits generated through different guiding criteria}
    \label{fig:verticalcell4}
\end{figure}

In this particular example, these ghost regions and the leverage of local outliers are not a problem for either method, but in real larger-dimensional datasets it could result in undesirable patterns if the sub-sample of data that a tree uses is not too representative. It might also pose a problem if a model is to be used for "novelty detection"  - that is, to identify anomalies in new data to which the model will not be fit, as opposed to identifying outliers within the same that that is available for producing a model. "Novelty detection" is however not the focus of this work.

In contrast to the gain criterion from \textsc{SCiForest} which is set on a random linear combination of a few variables, \textsc{DET} and related models such as \textsc{OCRF} and \textsc{GIF} will look at a larger part or even the entirety of the multivariate distribution of the data when evaluating a partition, and their criteria tend to favor partitions which divide the data more evenly in the earlier branches as opposed to leaving only a few points at one side of a division.

\textsc{DET}'s splitting criterion and scoring metric do not directly relate to the concept of isolation, and its intended use-case in \cite{det} was not exactly anomaly detection either, but as can be seen from their experiments and from the criterion that is maximized, the splits it generates will implicity isolate outlier observations faster (taking fewer splits) as they are assigned to higher-volume regions, which are in turn defined by fewer splits than smaller-volume regions. In \cite{ocrf} they found isolation depth to be a useful scoring metric after generating partitions by a density-based criterion, although the same density score from \textsc{DET} can also be used as an anomaly score (the lower, the more anomalous), thereby avoiding the issue of having to extrapolate a depth remainder.

Just like \textsc{SCiForest}, \textsc{DET} also tends to separate easily-divisible multimodal distributions quickly and to be relatively insensitive to the presence of local outliers in the data, but without suffering from the issue of ghost regions:
\begin{figure}[H]
\centerline{\includegraphics[totalheight=3.5cm]{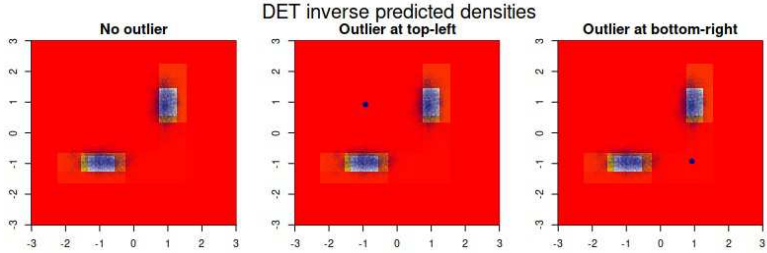}}
	\caption{Anomaly scores from \textsc{DET} in randomly-generated bimodal data.}
    \label{fig:verticalcell5}
\end{figure}

Unfortunately, it also tends to produce little discrimination in scores among non-extreme values, perhaps even less so than \textsc{SCiForest}. A logical thought to remedy this issue would be to employ a "forest" of \textsc{DET}s with sub-sampled observations just like \textsc{iForest} and without pruning the trees as originally proposed, but this does not eliminate the issue either:
\begin{figure}[H]
\centerline{\includegraphics[totalheight=3.5cm]{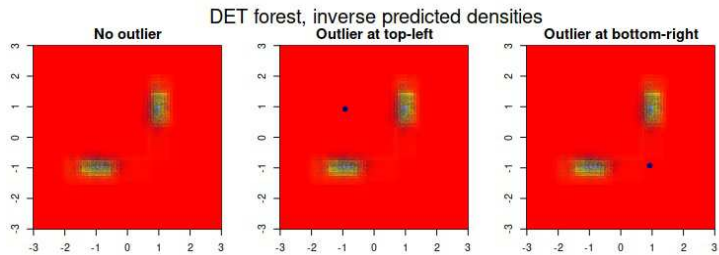}}
	\caption{Anomaly scores from an ensemble of \textsc{DET}s with subsampled data, in randomly-generated bimodal data.}
    \label{fig:verticalcell6}
\end{figure}

As another \textsc{iForest} variation, the \textsc{RRCF} model from \cite{rrcf} does not look at guiding the split threshold selection, but at guiding the selection of variable to split, doing so with a probability proportional to the range of each variable at a given node and relying on averaging out many randomized choices instead of always aiming for the best one as \textsc{DET} tries to do.

Such a criterion tries to address the issue of producing splits which are not helpful for discriminating outliers, but does so very differently from other methods. Variable selection is not a new idea for \textsc{iForest} - for example, in \cite{iso} they briefly mention using kurtosis as a variable screener, selecting only the top variables ranked by kurtosis to be included in a model, while \textsc{SCiForest} tries many variables at random in order to identify good candidates and \textsc{DET} tries all of them deterministically. \textsc{RRCF} is different in the sense that it relies on averaging out errors, introducing less-granular guidance to the split choices.

In theory, \textsc{RRCF} suffers from many of the same issues as \textsc{iForest} as outlined in e.g. \cite{ext} and \cite{inne}, and visualizing its calculated anomaly scores on randomly-generated points that exemplify an extreme pattern (such as inner and outer rings, sine waves, etc.) would not look too differently from the scores produced by \textsc{iForest} - for example, in the bimodal random numbers from earlier plots, it would not identify the top-left or bottom-right outliers as the most anomalous observations, although it would rank them as relatively more anomalous compared to the rest than \textsc{iForest} would. In practice however, \textsc{RRCF} has demonstrated improved performance metrics when applied on real datasets of varying characteristics.

\section{Better data separations}

Ideally, one would want the splits made by an isolation tree to assign more points to regions where the multivariate density or likelihood of the data distribution is higher and to make these regions as narrow as needed, or under an alternative view to group together observations which are more similar, making the points in a node closer and closer further down each tree, which in broad terms requires making splits on good boundaries across the right variables. This kind of problem has been approached from different angles in different problem domains such as anomaly detection, clustering, or approximate nearest neighbor search, using different measuring criteria. \textsc{DET} looked at volume, \textsc{RRCF} at ranges of variables, \textsc{SCiForest} at variance, \cite{iso} briefly looked at kurtosis, and given the logic of \textsc{iForest} one might also think of looking at other measurements of deviation from a uniform distribution. From all these criteria, the most natural (esp. for multimodal distributions) and most widely used in different algorithms is perhaps variance, but the way in which other algorithms use it is rather different from \textsc{SCiForest}.

Typically, supervised decision trees (e.g. \cite{c45} and \cite{cart}) make a trade-off in their guiding heuristic between branch purity (homonegenity of the points in a given partition) and number of observations that are assigned to each branch of a split. \textsc{DET} and similar also make such a trade-off implicitly, while \textsc{SCiForest} and \textsc{RRCF} do not look at this aspect.

More concretly, in the case of regression trees, it is typical to use as splitting criterion a pooled standard deviation gain, which is slightly different from the averaged standard deviation gain used by \textsc{SCiForest}
$$
\text{gain}_{\text{pooled}} = \frac{
	\sigma_{\text{all}}
	- \frac{  n_{\text{left}} \sigma_{\text{left}}
			+ n_{\text{right}} \sigma_{\text{right}} }{n_{\text{left}} + n_{\text{right}}}
}{\sigma_{\text{all}}}
$$

The "Fair-Cut Forest" or \textsc{FCF} from \cite{imputer} applied this pooled gain criterion on random linear combinations of variables in order to determine splits by these same linear combinations, similarly to \textsc{SCiForest} but with the goal of finding neighbors for missing value imputations rather than producing anomaly scores. While not the original goal of \textsc{FCF}, this split guiding criterion turns out to produce splits that are also useful for outlier detection as will be seen in the next sections, as these splits tend to represent more natural separations, which can come especially useful in clustered or multimodal distributions (as opposed to outliers represented by extreme values which could be identified by e.g. kurtosis or deviation w.r.t a uniform distribution):

\begin{figure}[H]
\centerline{\includegraphics[totalheight=4cm]{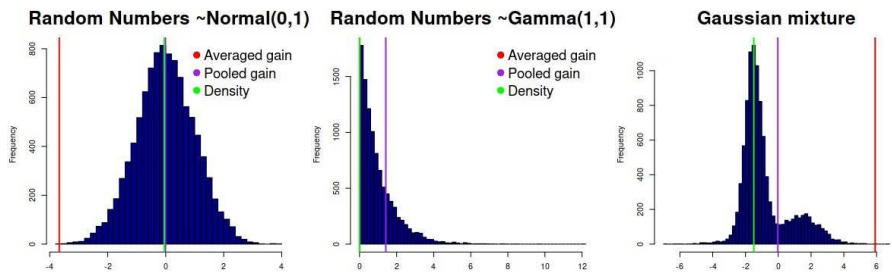}}
    \caption{Splits chosen by each criteria in randomly-generated numbers. In the left picture, the density-chosen point matches with that of pooled gain, while in the center picture it closely matches with that of averaged gain.}
    \label{fig:verticalcell7}
\end{figure}

If the best possible pooled gain from a split point is evaluated across different variables, those in which clusters are more easily formed or in which separations are more clear will result in a higher gain, perhaps making it also a potential heuristic for variable selection.

In a way, pooled gain applied on a single variable or linear combination is trying to "predict" the same values that it is input (using group means as "prediction"), and the groups that it produces will in expectation become more and more homogeneous after each split, resulting in something similar to an autoencoder; but it is a myopic criterion: if measuring a global squared error objective across all variables, "predictions" (per-column means in each node) are not guaranteed to improve after each step. One could also think of minimizing instead a pooled gain criterion calculated across all variables in the data, but such a criterion, while perhaps reasonable in theory, would in practice result in much slower evaluation and would perhaps lead to sub-optimal splits in the presence of multicollinearity. A shorter and myopic gain calculated on one or a few variables can nevertheless rely on averaging out errors across many runs for better results.

Compared to \textsc{SCiForest}, \textsc{FCF} can also be reasonably fast at isolating scattered outliers within each tree, but the partitions that it chooses to isolate them are rather different and can result in much larger variability in the final results according to the data sub-samples that each tree uses:
\begin{figure}[H]
\centerline{\includegraphics[totalheight=5.5cm]{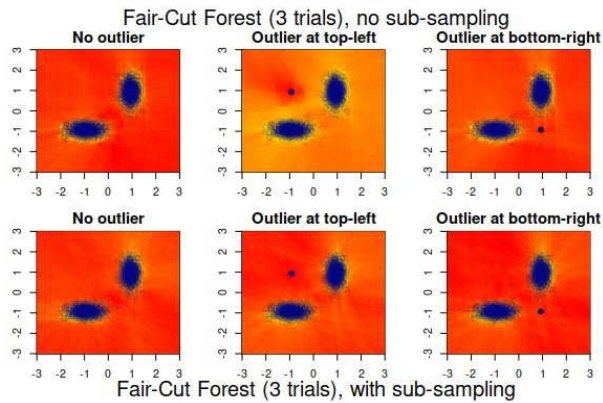}}
    \caption{Anomaly scores from \textsc{FCF} in randomly-generated bimodal data}
    \label{fig:verticalcell8}
\end{figure}

The tree structure produced by splits that follow this pooled gain criterion is also different from that of \textsc{iForest} - in the case of uniformly-random data, the optimal split would always correspond to assigning half of the points to one branch, leading to an expected isolation depth of $\mathbb{E}[d(m)] = log_2 m$ when $m$ is a power of 2, which is strictly smaller than the expected depth for uniformly random splits (given by $\mathbb{E}[d(m)] = 2 (H_m - 1)$, where $H_m$ is the m-th harmonic number) for $m > 2$, but with a structure that produces larger trees (containing more nodes) at the same height as the equivalent \textsc{iForest}:

\begin{figure}[H]
\centerline{\includegraphics[totalheight=3cm]{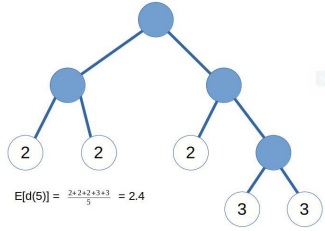}}
    \caption{Expected isolation depth for 5 uniform points under pooled gain}
    \label{fig:verticalcell9}
\end{figure}

\section{Hyperparameters under pooled gain}

In \cite{sci} they kept most of the same \textsc{iForest} hyperparameters for the proposed \textsc{SCiForest} algorithm, adding a randomized trial of many linear combinations on top. These same designs might not be the most optimal for \textsc{FCF} when used for anomaly detection however, and are rather different from what \textsc{DET} and \textsc{OCRF} used. In the case of \textsc{FCF}, it should be taken into consideration that:
\begin{itemize}
	\item If making splits by an averaged gain criterion, typically most of the cases will assign a single observation to one of the branches, and it could take many trials or evaluations across columns or linear combinations to choose a split that puts more than one observation in the smaller branch, which is not the case for the pooled gain criterion. What's more, if the split is performed on a linear combination of variables, it is typically enough for at least one of them to not be irrelevant in order for the linear combination to produce helpful splits.
	\item In \textsc{iForest} it was concluded that 100 trees was enough to get convergent results, but the splits generated by a pooled gain criterion on a sub-sample of the data have more expected variability, thus perhaps more trees would be needed to reach convergent results, especially in larger datasets.
	\item Both \textsc{iForest} and \textsc{SCiForest} used sub-sampling with 256 observations per tree, which in the case of \textsc{iForest} is helpful for dealing with some of the inherent problems in the method, but \textsc{FCF} might not suffer from the exact same problems.
	\item Both \textsc{iForest} and \textsc{SCiForest} chose a maximum depth equal to that of balanced-tree height as terminating criterion, but in the case of standardized pooled gain, it is also possible - and perhaps a more reasonable choice - to set this termination criterion on the gain itself, with nodes in which the best possible gain is less than some pre-determined acceptance level (e.g. 25\%) potentially being set as terminal nodes and a remainder depth extrapolated.
\end{itemize}

A small experiment with the "SpamBase" dataset would suggest that deeper trees are beneficial, regardless of the split criterion that is used:

\begin{table}[H]
\centering
\caption {Results on "SpamBase" dataset with different termination criteria}
\begin{adjustbox}{max width=\textwidth}{\centering
\begin{tabular}{|r|c|c|c|}
\hline
\textbf{Model} & \textbf{Termination} & \textbf{AUROC} & \textbf{Nodes} \\
 \hline
\textsc{iForest} & depth $\geq$ 8 & 0.6405 & 22,346 \\  \hline
\textsc{iForest} & depth $\geq$ 16 & 0.6545 & 60,514 \\  \hline
\textsc{iForest} & isolation & 0.6926 & 184,032 \\  \hline
\textsc{FCF} & depth $\geq$ 8 & 0.5897 & 44,100 \\  \hline
\textsc{FCF} & depth $\geq$ 16 & 0.6112 & 114,606 \\  \hline
\textsc{FCF} & gain $<$ 0.5 & 0.6197 & 132,694 \\  \hline
\textsc{FCF} & isolation & 0.6220 & 174,960 \\  \hline
\end{tabular}}\end{adjustbox}
\end{table}

Increasing the depth however also increases the standard error for the expected isolation depth that each observation would have:
\begin{figure}[H]
\centerline{\includegraphics[totalheight=6cm]{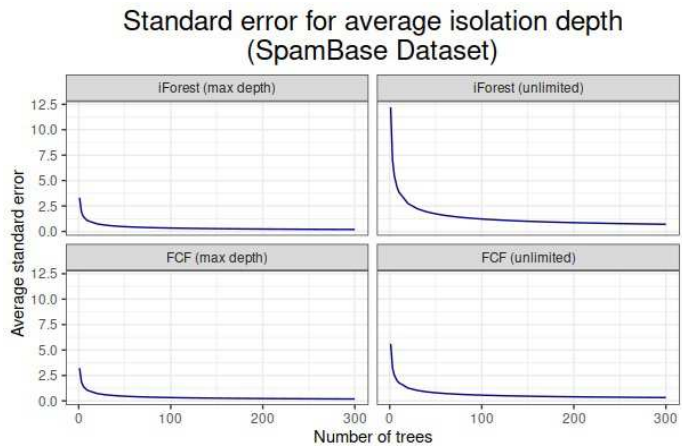}}
    \caption{Average standard error for expected isolation depth of each observation}
    \label{fig:verticalcell10}
\end{figure}

This increase in the standard error can nevertheless be easily offset by an increase in the number of isolation trees, at the expense of longer running times. For example, reaching the same average standard error per observation for the expected isolation depth as in \textsc{iForest} with 100 trees and limited depth under \textsc{FCF} with unlimited depth would require a bit over 200 trees in this particular dataset. Fitting such a model (trees growing until full isolation, twice as many trees) is roughly an order of magnitude slower than if using the hyperparameters proposed in \cite{iso}, but with a careful software implementation, this is still much faster than many competing methods for anomaly detection across several datasets as will be seen in the next section.

In \cite{sci} the suggestion was to make splits on random linear combinations of 2 variables, trying 10 such random linear combinations at each node. Some short experiments would suggest that 10 trials is perhaps too much for a pooled gain criterion, with better performance being achieved when this number is decreased, and only a small difference from using hyperplanes vs. axis-parallel splits: 
\begin{table}[H]
\centering
\caption {Varying the number of trials for different variables or random linear combinations of several variables in \textsc{FCF}, results on "SpamBase" dataset}
\begin{adjustbox}{max width=\textwidth}{\centering
\begin{tabular}{|r|r|c|c|c|}
\hline
\textbf{Variables per split} & \textbf{Trials} & \textbf{AUROC (256 samples)} & \textbf{AUROC (no subsampling)} \\
 \hline
1 & 1 & 0.7312 & 0.7095 \\  \hline
1 & all & 0.4405 & 0.4697* \\  \hline
2 & 1 & 0.7245 & 0.6950 \\  \hline
2 & 3 & 0.6220 & 0.5868 \\  \hline
2 & 10 & 0.4986 & 0.4861 \\  \hline
5 & 1 & 0.6893 & 0.6683 \\  \hline
5 & 3 & 0.6696 & 0.6199 \\  \hline
5 & 10 & 0.6127 & 0.5619 \\  \hline
\end{tabular}}\end{adjustbox}
\begin{tablenotes}
      \footnotesize \item * This case was sampled with replacement in order to avoid generating the exact same splits in each tree.
\end{tablenotes}
\end{table}

\begin{table}[H]
\centering
\caption {Varying the number of trials for different variables or random linear combinations of several variables in \textsc{FCF}, results on "Satellite" dataset}
\begin{adjustbox}{max width=\textwidth}{\centering
\begin{tabular}{|r|c|c|c|}
\hline
\textbf{Variables per split} & \textbf{Trials} & \textbf{AUROC (256 samples)} & \textbf{AUROC (no subsampling)} \\
 \hline
1 & 1 & 0.8368 & 0.7261 \\  \hline
1 & all & 0.8057 & 0.7152* \\  \hline
2 & 1 & 0.8186 & 0.7465 \\  \hline
2 & 3 & 0.8414 & 0.7451 \\  \hline
2 & 10 & 0.8298 & 0.7312 \\  \hline
5 & 1 & 0.8246 & 0.7544 \\  \hline
5 & 3 & 0.8235 & 0.7618 \\  \hline
5 & 10 & 0.8080 & 0.7509 \\  \hline
\end{tabular}}\end{adjustbox}
\begin{tablenotes}
      \footnotesize \item * This case was sampled with replacement in order to avoid generating the exact same splits in each tree.
\end{tablenotes}
\end{table}

The number of samples to take for each tree has a similar influence as in \textsc{iForest}, but a deeper look at some datasets would reveal that the optimal sample size for some datasets can be much smaller than the 256 proposed in \cite{iso} which was also used in others such as \cite{ext} or \cite{sci}:

\begin{figure}[H]
\centerline{\includegraphics[totalheight=6cm]{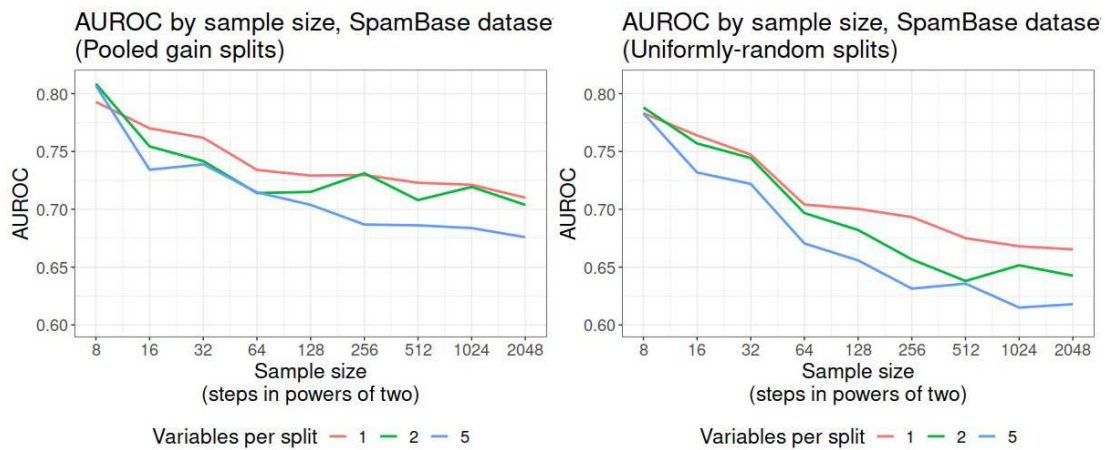}}
    \caption{AUROC by sample size and model characteristics in "SpamBase" dataset}
    \label{fig:verticalcell11}
\end{figure}

Yet in other datasets, sub-sampling is not helpful, with trees fitted to the full number of rows achieving better performance:

\begin{figure}[H]
\centerline{\includegraphics[totalheight=6cm]{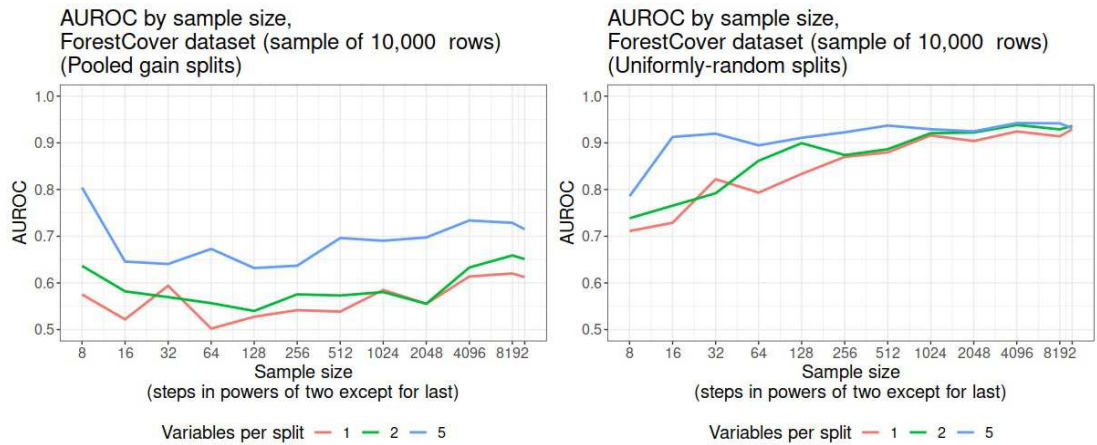}}
    \caption{AUROC by sample size and model characteristics in sub-sampled "ForestCover" dataset}
    \label{fig:verticalcell12}
\end{figure}

As such, a potentially safe combination of hyperparameters for using \textsc{FCF} in anomaly detection would be as follows:
\begin{itemize}
	\item Trees grown until full isolation (only one point is present in a given branch of a split) or until no further split is possible, in which case a remainder is extrapolated.
	\item 200 trees.
	\item A sample size of 32 to 256 points chosen without replacement to be used by each tree, but with numbers much smaller or much larger than this resulting in better results in some datasets (perhaps smaller numbers being better for multi-modal datasets).
	\item Splitting data according to a threshold on a random linear combination of 2 variables, but with a larger number of variables and with single-variable (axis-parallel) splits producing better results in some datasets.
	\item Only one trial of a uniformly-randomly-chosen candidate variable or random linear combination to try at each node.
\end{itemize}

The full procedure for producing trees and anomaly scores is outlined below. The implementation produced here is made open source and freely available\footnote{\url{https://www.github.com/david-cortes/isotree}}.

\begin{algorithm}[H]
\caption{FairCutTree}\label{FCT}
\hspace*{\algorithmicindent}
	\textbf{Inputs}
		Input data points $\mathbf{X}_{m \times n}$,
		number of splitting variables $p$,
		current depth $d$
\begin{algorithmic}[1]
\If {$m = 1$}
	\State Set as terminal node with depth $d$
\EndIf
\If {each of the $n$ columns in $\mathbf{X}$ has the same value across all of its rows}
	\State Set as terminal node with depth $d + \mathbb{E}[\text{depth}(m)]$
\EndIf
\State Initialize vector $\mathbf{z} := \textbf{0}^m$
\For {$1..p$}
	\State Choose a variable $v$ uniformly at random from $[1, n]$ such that it contains more than 1 unique value within $\mathbf{X}$, and define $\mathbf{y} = \mathbf{X}[:, v]$ (vector with the values of $\mathbf{X}$ in that variable)
	\State Draw a random coefficient $c \sim \text{Normal}(0,1)$
	\State Update $\mathbf{z} := \mathbf{z} + c \frac{\mathbf{y} - \bar{\mathbf{y}}}{\sigma_{\mathbf{y}}}$
\EndFor
\State Find the point $s$ that minimizes
$$
\frac{m_{\text{left}} \sigma_{\mathbf{z}_l} + m_{\text{right}} \sigma_{\mathbf{z}_r}}{m_{\text{left}} + m_{\text{right}}}
$$
Where $\mathbf{z}_l = \{ z \in \mathbf{z} \:\:\: | \:\:\: z \leq s \}$, $\mathbf{z}_r = \{ z \in \mathbf{z} \:\:\: | \:\:\: z > s \}$, and $\sigma_{\textbf{z}}$ is the standard deviation of $\mathbf{z}$
\State Define $\mathbf{X}_l = \{ \mathbf{x}_i \in \mathbf{X} \:\:\: | \:\:\: z_i \leq s \}$ and $\mathbf{X}_r = \{ \mathbf{x}_i \in \mathbf{X} \:\:\: | \:\:\: z_i > s \}$
\State Set $\text{Node}_L = \text{FairCutTree}(\mathbf{X}_l, p, d+1)$ and $\text{Node}_R = \text{FairCutTree}(\mathbf{X}_r, p, d+1)$
\end{algorithmic}
\end{algorithm}

\begin{algorithm}[H]
\caption{Fair-Cut Forest}\label{FCC}
\hspace*{\algorithmicindent}
	\textbf{Inputs}
		Input data points $\mathbf{X}_{m \times n}$,
		number of splitting variables $p$,
		number of trees $t$,
		sample size $s$
\begin{algorithmic}[1]
\For {$1..t$}
	\State Choose a set $\mathbb{S}$ of $s$ points uniformly at random from $[1,m]$ without replacement, and set $\mathbf{X}_s = \{ \mathbf{x}_i \in \mathbf{X} \:\:\: | \:\:\: i \in \mathbb{S} \}$
	\State Set $\text{Tree}_i = \text{FairCutTree}(\mathbf{X}_s, p, 0)$
\EndFor
\State Calculate $q = \mathbb{E}[\text{depth}(m)]$
\State \Return The generated set $\mathbb{T}$ of $t$ trees and $q$.
\end{algorithmic}
\end{algorithm}

\begin{algorithm}[H]
\caption{TreeScore}\label{TreeScore}
\hspace*{\algorithmicindent}
	\textbf{Inputs}
		Input point vector $\mathbf{x}_n$,
		tree node $\mathbb{N}$
\begin{algorithmic}[1]
\If {$\mathbb{N}$ is a terminal node}
	\State \Return The associated depth $d$ from $\mathbb{N}$
\Else
	\State initialize $z := 0$
	\For {$v = 1..p$ from the variables $v \in \mathbb{N}$}
		\State Update $z := z + c_v \frac{x_v - \bar{\mathbf{y}_v}}{\sigma_{\mathbf{y}_v}}$ with the $c_v$, $\bar{\textbf{y}_v}$ and $\sigma_{\textbf{y}_v}$ that were determined for variable $v$ in $\mathbb{N}$
	\EndFor
	
	\If {$z \leq s$ from the optimal $s$ in $\mathbb{N}$}
		\State \Return $\text{TreeScore}(\mathbf{x}, \text{Node}_L)$ from the nodes in $\mathbb{N}$
	\Else
		\State \Return $\text{TreeScore}(\mathbf{x}, \text{Node}_R)$ from the nodes in $\mathbb{N}$
	\EndIf

\EndIf
\end{algorithmic}
\end{algorithm}

\begin{algorithm}[H]
\caption{ScorePoint}\label{ScorePoint}
\hspace*{\algorithmicindent}
	\textbf{Inputs}
		Input point vector $\mathbf{x}_n$,
		forest of $t$ fair-cut trees $\mathbb{T}$,
		expected isolation depth $q$
\begin{algorithmic}[1]
\State Initialize $d = 0$
\For {$i = 1..t$}
	\State Update $d := d + \text{ScoreTree}(\mathbf{x}, \text{Node}_0 \:\:\:\text{from}\:\:\: \mathbb{T}_t)$
\EndFor
\State \Return $2^{- \frac{d / t}{q}}$
\end{algorithmic}
\end{algorithm}

\section{Evaluating different methods}

The variation of the "Fair-Cut forest" algorithm or \textsc{FCF} proposed here was compared against \textsc{iForest}, some of its variations, and other competing methods for anomaly detection, including:
\begin{itemize}

\item The original "isolation forest" or \textsc{iForest} from \cite{iso}, reimplemented according to the paper and sharing the same codebase as the \textsc{FCF} here. The hyperparameters used were the ones suggested in \cite{iso} (100 trees, 256 samples per tree, maximum depth of 8), and most of the other algorithms are based around these same hyperparameters.

\item The original \textsc{iForest} but with the same hyperparameters set for \textsc{FCF} (200 trees, unlimited depth, marked as \textsc{IF-u}).

\item The "extended isolation forest" or \textsc{EIF} from \cite{ext}, using both the authors' implementation\footnote{\url{https://github.com/sahandha/eif}} (\textsc{EIF-o}) and and independent implementation (\textsc{EIF-t}). Note that the implementation here differs from the original \textsc{EIF} in one important aspect: it standardizes variables by calculating the standard deviations at each node rather than just at the beginning of the tree. The extension level was set to 1 (splitting by 2 variables at a time) as recommended by the authors, plus the same default hyperparameters from \textsc{iForest}.

\item The "robust random-cut forest" or \textsc{RRCF} from \cite{rrcf}, using a publicy-available implementation\footnote{\url{https://github.com/kLabUM/rrcf}} that uses the "CoDisplacement" metric as proposed in \cite{rrcf} for anomaly scoring (denoted as \textsc{RRCF-c}), and a different public implementation\footnote{\url{https://github.com/navdeep-G/robust-random-cut-forest}} that uses average isolation depth for anomaly scoring (denoted as \textsc{RRCF-d}). Note that both of them are pure-Python implementations, which makes them much slower than the rest despite the idea being in theory relatively fast to execute, and that the second implementation did not allow setting random seeds.

\item The "density estimation trees" or \textsc{DET} from \cite{det}, using the implementation from MLPack\footnote{\url{https://www.mlpack.org/doc/mlpack-3.0.4/doxygen/dettutorial.html}}.

\item A "forest" (\textsc{DEF}) of 100 \textsc{DET}s using sub-samples of 256 points each, sampled without replacement, and the trees built without pruning, in order to make it directly comparable to \textsc{iForest}, with the final prediction averaged across these 100 \textsc{DET}s.

\item The "generalized isolation forest" or \textsc{GIF} from \cite{genif}, using the implementation provided by the authors\footnote{\url{https://github.com/philippjh/genif}} and the hyperparameters suggested in their web example.

\item The "one-class random forest" or \textsc{OCRF} from \cite{ocrf}, using the implementation provided by the authors\footnote{\url{https://github.com/ngoix/OCRF}}.

\item The \textsc{SCiForest} from \cite{sci}, reimplemented according to their description (including e.g. range penalties at prediction time) and sharing the same codebase as \textsc{FCF}. Another variation with 200 trees grown until full isolation of every point was also explored (\textsc{SCiF-u}), driven by the expected isolation depth issue highlighted earlier. Note that the original \textsc{iForest} code from the authors\footnote{\url{https://sourceforge.net/projects/iforest/}} (of both \textsc{IF} and \textsc{SCiForest}) included a similar decision criterion but it used only single-variable splits, and was thus not compared against here.

\item The "isolation nearest-neighbor ensembles" or \textsc{iNNE} from \cite{inne}, using one of the implementations from the authors\footnote{\url{https://github.com/zhuye88/iNNE}}, with hyperparameters $\psi=32$ and $t=100$ as suggested in the reference. Note that this implementation did not allow setting random seeds and was ran only once.
\item The "local outlier factor" or \textsc{LOF} from \cite{lof}, using the implementation from scikit-learn\footnote{\url{https://scikit-learn.org/stable/modules/generated/sklearn.neighbors.LocalOutlierFactor.html}}.
\item The "one-class support vector machine" or \textsc{OCSVM} from \cite{ocsvm} in its linear and RBF kernel versions, using the implementation from scikit-learn\footnote{\url{https://scikit-learn.org/stable/modules/generated/sklearn.svm.OneClassSVM.html}}.
\end{itemize}

Models are evaluated on public datasets downloaded from the ODDS repository\footnote{\url{http://odds.cs.stonybrook.edu}} and from \cite{repo2}\footnote{\url{https://www.dbs.ifi.lmu.de/research/outlier-evaluation/}}, representing a variety of outlier types (e.g. minority-class, extreme-valued, minority-mode) and problem domains. All datasets are taken in their raw form (which differs from how some of the references use them), prefering the version from ODDS when available. The evaluation is done by fitting said models to the full data under their default hyperparameters and producing outlier scores on this same data. These scores are compared against the true labels for outlierness (which the models do not observe in their fitting procedures) in terms of area under the receiver-operating characteristic curve (\textsc{ROC}) and area under the precision-recall curve (\textsc{PR}), which as analyzed in \cite{aupr} might be a more appropriate metric than the more commonly used \textsc{ROC}. Each model that uses randomization is fitted 10 times using different random seeds, and the result reported here is calculated as the mean across these 10 runs. For larger datasets, some of the methods were not compared against due to the long running times that some of them would require as the volume of data grows.

Experiments were run on an AMD Ryzen 7 2700 CPU with 8 cores running at 3.2GHz, and times are reported in seconds taken for fitting a given model plus producing outlier scores from said model. Where possible, software was compiled under GCC version 10.3 with options "-O3" and "-march=native". Note however that, for larger datasets, the times for tree-based methods are dominated by the calculation of outlier scores, which was done with the same data format as the models (column-major order, double precision) and through the same software library that produced each model, but these times could be reduced significantly if, for example, a row-major format were used and the decision trees were pre-compiled (e.g. through software such as "treelite"\footnote{\url{https://treelite.readthedocs.io/en/latest/index.html}}), which would make all tree-based methods look significantly faster (roughly 10 to 30 times faster in the larger datasets if the compilation times are not considered).

While others such as \cite{genif} and \cite{inne} perform hyperparameter tuning in their experiments based on the true outlier labels, in the case of anomaly detection, this is a problematic aspect as such labels are not supposed to be available in contexts in which models like these would be employed, and it can tend to favor model types which are only good under fine-tuned parameters specific to each dataset without providing a clear picture of how generalizable such models really are. For example, models such as \textsc{GIF} or \textsc{iNNE} can show either one of the best or one of the worst performances in a given dataset depending on the hyperparameters used, while models such as \textsc{IF} or \textsc{EIF} are quite generalizable across datasets and outlier types under the same set of parameters.

\begin{table}[H]
\centering
\caption {Datasets used for comparisons}
\begin{adjustbox}{max width=\textwidth}{\centering
\begin{tabular}{|r|c|c|c|c|}
 \hline
 \textbf{Dataset} & \textbf{Rows} & \textbf{Columns} & \textbf{Outliers} \\
 \hline
Arrhythmia & 452 & 274 & 15\% \\  \hline
Pima & 768 & 8 & 35\% \\  \hline
SpamBase & 4,601 & 57 & 39.4\% \\  \hline
Satellite & 6,435 & 36 & 32\% \\  \hline
Pendigits & 6,870 & 16 & 2.27\% \\  \hline
Annthyroid & 7,200 & 6 & 7.42\% \\  \hline
Mnist & 7,603 & 100 & 9.2\% \\  \hline
ALOI & 50,000 & 27* & 3\% \\  \hline
ForestCover & 286,048 & 10 & 0.9\% \\  \hline
\end{tabular}}\end{adjustbox}
\begin{tablenotes}
      \footnotesize \item * This dataset contains a categorical variable. As most of the methods compared here do not work with categorical variables, it was left out for a fairer comparison.
\end{tablenotes}
\end{table}

Some additional comments about these datasets:\itemize{
 \item Arrhythmia: variables are a good mixture of integral, continuous, binary, near-symmetric, and power tail. This was originally a multi-class classification dataset about different types of cardiac arrythmia, in which the non-arrythmic cases were mixed together with the most common arrythmias and the rest set as outliers.
 \item Pima: this was originally a binary-classification dataset in which the minority class was set as outliers.
 \item SpamBase: many variables have a skewed distribution with the majority of the rows having a value of exactly zero.
 \item Satellite: all variables have integer-only values, which can lead to earlier termination of tree-based methods due to ending with non-unique values in a node. This dataset was originally meant for multi-class classification, and was adapted for anomaly detection by mixing together frequent classes and uncommon classes, making it in a way a more multi-modal dataset in which outliers are those near less common modes, which is quite different from the other datasets.
 \item Annthyroid: most outliers can be differentiated through a single high-skew column. This dataset originally consisted of 3 classes which can be thought of as representing values "low", "normal", "high"; with "low" and "high" set as outliers (that is, outliers are a mixture of opposite groups, representing a relatively different scenario as in most other datasets).
 \item Mnist: variables are non-integer but with relatively few unique values, also leading to potential earlier termination. This dataset was originally about multi-class classification, and was converted to anomaly detection by including a majority of rows from one class and a minority of rows from another class (all outliers are the same class).
 \item ALOI: contains many very low-variance columns in which most of the points have the same value. Also contains some binary and power-tailed variables. This is a dataset about images from different physical objects in which outliers constitute "rare objects with 1-10 instances" (\cite{aloi2}).
  \item ForestCover: just like "Mnist", this was originally a multi-class classification dataset that was converted by sampling from a majority and a minority class.
}

\begin{table}[H]
\centering
\caption {Results obtained by each method, part 1}
\begin{adjustbox}{max width=\textwidth}{\centering
\begin{tabular}{|r|c|c|c|c|c|c|c|c|c|}
 \hline
 &   \multicolumn{3}{|c|}{ \textbf{Arrythmia}}
 &   \multicolumn{3}{|c|}{ \textbf{Pima}}
 &   \multicolumn{3}{|c|}{ \textbf{SpamBase}}
  \\
 \textbf{Model} &
   \textbf{ROC} & \textbf{PR} & \textbf{Time} &
   \textbf{ROC} & \textbf{PR} & \textbf{Time} &
   \textbf{ROC} & \textbf{PR} & \textbf{Time}
 \\ \hline
\textsc{IF} &
	0.7938 & 0.4599 & \textbf{0.00408} &
	0.6795 & 0.4972 & \textbf{0.00537} &
	0.6362 & 0.4757 & \textbf{0.00774}
	\\ \hline
\textsc{IF-u} &
	0.804 & 0.5012 & 0.0103 &
	0.6807 & 0.5029 & 0.0119 &
	0.6927 & 0.5242 & 0.0294
	\\ \hline
\textsc{EIF-o} &
	0.8052 & 0.4707 & 0.798 &
	0.6628 & 0.5003 & 0.0784 &
	0.6229 & 0.4791 & 0.61
	\\ \hline
\textsc{EIF-t} &
	0.7986 & 0.4669 & 0.00735 &
	0.6778 & 0.4970 & 0.00633 &
	0.6596 & 0.5021 & 0.0149
	\\ \hline
\textsc{SCiF} &
	0.6854 & 0.3368 & 0.0188 &
	0.6130 & 0.4236 & 0.0172 &
	0.4517 & 0.3801 & 0.0189
	\\ \hline
\textsc{SCiF-u} &
	0.7333 & 0.3431 & 0.152 &
	0.6565 & 0.4548 & 0.122 &
	0.4223 & 0.3454 & 0.878
	\\ \hline
\textsc{FCF} &
	0.8032 & 0.4842 & 0.0411 & 
	\textbf{0.7362} & \textbf{0.5508} & 0.0363 & 
	\textbf{0.7321} & 0.5694 & 0.281
	\\ \hline
\textsc{RRCF-d} &
	\textbf{0.8134} & 0.5177 & 10.5 &
	0.6277 & 0.4809 & 12.1 &
	0.7255 & \textbf{0.6317} & 65
	\\ \hline
\textsc{RRCF-c} &
	0.7874 & 0.4279 & 3.96 &
	0.5900 & 0.4307 & 3.09 &
	0.5774 & 0.4924 & 4.08
	\\ \hline
\textsc{DET} &
	0.6706 & 0.3496 & 0.0979 &
	0.6042 & 0.4360 & 0.0321 &
	0.3862 & 0.4391 & 0.415
	\\ \hline
\textsc{DEF} &
	0.6845 & 0.3752 & 0.97 &
	0.6081 & 0.4940 & 0.212 &
	0.5999 & 0.4399 & 0.742
	\\ \hline
\textsc{OCRF} &
	NA* & NA* & 0.286 &
	0.6644 & 0.4743 & 0.295 &
	0.4335 & 0.3455 & 0.602
	\\ \hline
\textsc{GIF} &
	0.8031 & \textbf{0.5184} & 1.47 &
	0.6067 & 0.4627 & 0.076 &
	0.6580 & 0.5451 & 0.217
	\\ \hline
\textsc{iNNE} &
	0.6881 & 0.2776 & 0.1602 &
	0.5947 & 0.4155 & 0.0687 &
	0.4769 & 0.3698 & 0.1938
	\\ \hline
\textsc{ocSVM-L} &
	0.5498 & 0.3144 & 0.02 &
	0.2598 & 0.2417 & 0.0133 &
	0.2349 & 0.2658 & 1.17
	\\ \hline
\textsc{ocSVM-K} &
	0.7948 & 0.4843 & 0.0378 &
	0.6580 & 0.4907 & 0.0275 &
	0.5063 & 0.5062 & 1.75
	\\ \hline
\textsc{LOF} &
	0.7891 & 0.4189 & 0.0322 &
	0.5424 & 0.3727 & 0.00741 &
	0.4597 & 0.3576 & 0.525
	\\ \hline
\end{tabular}}\end{adjustbox}
\begin{tablenotes}
      \footnotesize \item * \textsc{OCRF} in some cases ended up producing the same outlier score for every single point, thus metrics could not be computed.
\end{tablenotes}
\end{table}

\begin{table}[H]
\centering
\caption {Results obtained by each method, part 2}
\begin{adjustbox}{max width=\textwidth}{\centering
\begin{tabular}{|r|c|c|c|c|c|c|c|c|c|}
 \hline
 &   \multicolumn{3}{|c|}{ \textbf{Satellite}}
 &   \multicolumn{3}{|c|}{ \textbf{Pendigits}}
 &   \multicolumn{3}{|c|}{ \textbf{Annthyroid}}
  \\
 \textbf{Model} &
   \textbf{ROC} & \textbf{PR} & \textbf{Time} &
   \textbf{ROC} & \textbf{PR} & \textbf{Time} &
   \textbf{ROC} & \textbf{PR} & \textbf{Time}
 \\ \hline
\textsc{IF} &
	0.7164 & 0.6624 & \textbf{0.013} &
	0.9549 & 0.2268 & \textbf{0.0151} &
	0.8300 & 0.2820	& 0.0126
	\\ \hline
\textsc{IF-u} &
	0.7280 & 0.6431 & 0.0382 &
	0.9635 & 0.3029 & 0.04 &
	0.8385 & 0.3121	& 0.0518
	\\ \hline
\textsc{EIF-o} &
	0.6970 & 0.6849 & 0.808 &
	0.9736 & 0.3616 & 0.536 &
	0.7407 & 0.2675 & 0.393
	\\ \hline
\textsc{EIF-t} &
	0.6800 & 0.6364 & 0.017 &
	0.9592 & 0.3391 & 0.022 &
	0.8207 & 0.2859 & 0.021
	\\ \hline
\textsc{SCiF} &
	0.6083 & 0.5888 & 0.0285 &
	\textbf{0.9788} & \textbf{0.4110} & 0.0337 &
	0.7715 & 0.4634 & 0.0267
	\\ \hline
\textsc{SCiF-u} &
	0.7549 & 0.6916 & 0.356 &
	0.9785 & 0.3552 & 0.209 &
	0.9668 & 0.6594 & 0.721
	\\ \hline
\textsc{FCF} &
	\textbf{0.8253} & \textbf{0.7300} & 0.139 & 
	0.9662 & 0.2730 & 0.179 & 
	0.8712 & 0.3810 & 0.203
	\\ \hline
\textsc{RRCF-d} &
	0.7270 & 0.6837 & 69 &
	0.9374 & 0.2310 & 65 &
	0.7109 & 0.2271 & 81
	\\ \hline
\textsc{RRCF-c} &
	0.7089 & 0.5747 & 4.43 &
	0.9198 & 0.2125 & 3.41 &
	0.7477 & 0.2457 & 3.43
	\\ \hline
\textsc{DET} &
	0.7070 & 0.5495 & 1.27 &
	0.7088 & 0.0384 & 0.558 &
	0.8910 & 0.2687 & 0.825
	\\ \hline
\textsc{DEF} &
	0.7300 & 0.4644 & 0.973 &
	0.8602 & 0.2583 & 0.517 &
	\textbf{0.9789} & \textbf{0.7147} & 0.612
	\\ \hline
\textsc{OCRF} &
	0.7495 & 0.6568 & 0.711 &
	0.8833 & 0.1667 & 0.776 &
	0.8368 & 0.2094 & 0.613
	\\ \hline
\textsc{GIF} &
	0.7561 & 0.5485 & 0.218 &
	0.8501 & 0.0907 & 0.134 &
	0.5342 & 0.0885 & 0.0679
	\\ \hline
\textsc{iNNE} &
	0.6042 & 0.3892 & 0.189 &
	0.8975 & 0.1064 & 0.1265 &
	0.7047 & 0.1891 & 0.1309
	\\ \hline
\textsc{ocSVM-L} &
	0.6759 & 0.4803 & 1.47 &
	0.7652 & 0.1233 & 1.28 &
	0.4857 & 0.0840 & \textbf{0.00125}
	\\ \hline
\textsc{ocSVM-K} &
	0.6669 & 0.6718 & 2.62 &
	0.9599 & 0.3206 & 2.39 &
	0.5727 & 0.1171 & 2.35
	\\ \hline
\textsc{LOF} &
	0.5428 & 0.3754 & 0.784 &
	0.4821 & 0.0368 & 0.972 &
	0.7373 & 0.2055 & 0.0773
	\\ \hline
\end{tabular}}\end{adjustbox}
\end{table}

\begin{table}[H]
\centering
\caption {Results obtained by each method, part 3}
\begin{adjustbox}{max width=\textwidth}{\centering
\begin{tabular}{|r|c|c|c|c|c|c|c|c|c|}
 \hline
 &   \multicolumn{3}{|c|}{ \textbf{Mnist}}
 &   \multicolumn{3}{|c|}{ \textbf{ALOI}}
 &   \multicolumn{3}{|c|}{ \textbf{ForestCover}}
  \\
 \textbf{Model} &
   \textbf{ROC} & \textbf{PR} & \textbf{Time} &
   \textbf{ROC} & \textbf{PR} & \textbf{Time} &
   \textbf{ROC} & \textbf{PR} & \textbf{Time}
 \\ \hline
\textsc{IF} &
	0.8033 & 0.2608 & \textbf{0.0147} & 
	0.5375 & 0.0327 & \textbf{0.029} & 
	0.8509 & 0.0379 & \textbf{0.222}
	\\ \hline
\textsc{IF-u} &
	0.8167 & 0.2851 & 0.0474 & 
	0.5291 & 0.0319 & 0.211 & 
	0.8552 & 0.0388 & 0.935
	\\ \hline
\textsc{EIF-o} &
	0.8150 & 0.2669 & 1.81 & 
	0.5428 & 0.0351 & 4.01 & 
	0.9047 & 0.0634 & 16.7
	\\ \hline
\textsc{EIF-t} &
	0.8130 & 0.2762 & 0.0281 & 
	0.5428 & 0.0332 & 0.0755 & 
	0.8720 & 0.0445 & 0.48
	\\ \hline
\textsc{SCiF} &
	0.8584 & 0.3864 & 0.0285 & 
	0.5192 & 0.0306 & 0.0698 & 
	0.7037 & 0.0154 & 0.405
	\\ \hline
\textsc{SCiF-u} &
	0.8432 & 0.3762 & 0.316 &
	0.5173 & 0.0317 & 7.31 &
	0.7040 & 0.0157 & 11.3
	\\ \hline
\textsc{FCF} &
	0.7871 & 0.2532 & 0.228 & 
	0.5256 & 0.0311 & 2.06 & 
	0.5463 & 0.0107 & 3.58
	\\ \hline
\textsc{RRCF-d} &
	0.8204 & 0.2944 & 76 & 
	0.5484 & 0.0434 & 614 &
	0.7269 & 0.0158 & 2393
	\\ \hline
\textsc{RRCF-c} &
	0.7955 & 0.3026 & 4.45 & 
	0.5659 & 0.0441 & 20.1 &
	0.8994 & 0.0465 & 77
	\\ \hline
\textsc{DET} &
	0.8087 & 0.2645 & 1.37 & 
	0.5001 & 0.0317 & 0.0389 & 
	0.6679 & 0.0166 & 1256
	\\ \hline
\textsc{DEF} &
	0.6390 & 0.2005 & 0.986 & 
	0.4868 & 0.0320 & 2.91 & 
	0.5875 & 0.0178 & 15.5
	\\ \hline
\textsc{OCRF} &
	NA* & NA* & 0.756 & 
	0.5026 & 0.0310 & 2.44 & 
	0.8344 & 0.0419 & 24
	\\ \hline
\textsc{GIF} &
	0.7934 & 0.2412 & 0.642 & 
	0.5297 & 0.0406 & 0.244 & 
	0.7051 & 0.0281 & 0.796
	\\ \hline
\textsc{iNNE} &
	0.8324 & 0.3641 & 0.4657 & 
	\textbf{0.5814} & \textbf{0.0477} & 0.8570 & 
	0.9584 & 0.1871 & 2.2602
	\\ \hline
\textsc{ocSVM-L} &
	\textbf{0.9449} & \textbf{0.6709} & 3.45 & 
	0.5255 & 0.0412 & 148 &
	\textbf{0.9824} & \textbf{0.3904} & 3670
	\\ \hline
\textsc{ocSVM-K} &
	0.8216 & 0.3231 & 5.4 & 
	0.5172 & 0.0460 & 162.63 &
	0.6565 & 0.0125 & 4316 
	\\ \hline
\textsc{LOF} &
	0.6482 & 0.1874 & 1.33 & 
	- & - & - &
	- & - & -
	\\ \hline
\end{tabular}}\end{adjustbox}
\end{table}

\section{Discussion}

As expected, there was no dominant outlier detection model across the datasets experimented with here, which consitute a mixture of different oulier types. Taking \textsc{IF} as a base, the only variation that sees consistently better or not-much-worse results across all types of datasets and outliers is \textsc{EIF}, at least when examined under the default hyperparameters of each model, which one might prefer to use in the absence of labels. Making the trees deeper and the forest larger also results in a small performance improvement at the cost of much longer running times, but these times are still competitive against other methods.

When these datasets are viewed in terms of outlier types however, one would notice that clustered outliers from multimodal datasets (e.g. "Arrythmia", "Satellite", "SpamBase", "Annthyroid") - oftentimes those of utmost interest but also the hardest to flag - are better identified under tree-based models than under other families, with non-uniformly-random splits providing an edge. In particular, the \textsc{FCF} model with hyperparameters chosen more appropriately for anomaly detection (as compared to its original application field) looks to be the best performer in these cases, followed closely by \textsc{RRCF-d} (which uses isolation depth for calculating anomaly scores instead of the "co-displacement" metric initially proposed for this model), but with a rather large advantage over \textsc{RRCF} in terms of computational efficiency (should be noted again that the \textsc{RRCF} implementation used here was not as optimized, but the procedure itself requires checking values across all columns of the data at each node, which makes it less scalable than other split guiding heuristics). For other types of outliers such as those of minority-in-binary-classes however (e.g. "Mnist", "ForestCover"), \textsc{FCF} sees degraded performance compared to \textsc{IF}, and other families of less-local methods present better performance than either of them.

\textsc{SCiForest} was found in \cite{sci} to provide an increase in AUROC under the majority of the datasets experimented with. The experiments here however show a very different picture once datasets with more varied characteristics are evaluated - for example, in the "SpamBase" dataset, \textsc{SCiForest} results in a very large performance drop compared to \textsc{IF}, even driving the AUROC below 0.5 (which is what random predictions would achieve). It should also be noted that these experiments failed to reproduce similar AUROC numbers under the same datasets used in \cite{sci}, but once the depth limit was removed from the algorithm, the AUROC metrics became very similar to those reported there.

In \cite{ocrf}, they report AUROC and AUPR metrics for many of the same datasets and baselines as in here, following a very similar methodology, but the results obtained in this experiment differ very significantly from those reported in \cite{ocrf}, both for the proposed \textsc{OCRF} algorithm as well as for the baselines compared against - for example, \cite{ocrf} reports an AUROC of 0.850 in the SpamBase dataset for \textsc{OCRF}, while the experiments here reached an average AUROC of 0.4335 only, and while \cite{ocrf} found \textsc{OCRF} to outperform \textsc{IF} in 10 out of 12 datasets, that was the case only in 2 out of the 9 datasets experimented with in here (down to 1 dataset if considering \textsc{IF-u}). What's more, the proposed \textsc{OCRF} algorithm in some cases ended up producing the exact same outlier score for all observations, which made calculation of metrics not meaningful.

Computation times in this experiment also showed how different implementations of the same algorithm can result in widly different execution speed, suggesting that the running times reported by previous works might not provide fair comparisons of the relative efficiency of algorithms. For example, \cite{genif} produced a very efficient software implementation of the proposed \textsc{GIF} algorithm, reporting \textsc{GIF} to run around an order of magnitude faster than \textsc{EIF} on the "SpamBase" dataset, and the experiments here could indeed confirm that it ran significantly faster than the \textsc{EIF} provided by its own authors, but once \textsc{EIF} is reimplemented with more attention to speed concerns, the comparison then flips as the \textsc{EIF} used here was an order of magnitude faster than \textsc{GIF}, despite doing extra computations at each step compared to the original \textsc{EIF} (when standardizing variables). Although not shown here, it was possible to obtain a further order of magnitude speed-up for prediction times in the larger datasets for models that use axis-parallel splits through pre-compilation of the trees (through e.g. the "treelite" software), but most of the software implementations compared here did not offer such functionality.

\section{Conclusions}

This work analyzed different heuristics that have been proposed to guide the choice of splits in isolation forests, highlighting some behaviors and properties of interest of the models obtained by following different criteria; and along the way proposed a splitting rule based on uniformly-random column choice with deterministic split threshold selection which is obtained by maximizing a pooled information gain metric, aimed at increasing outlier detection capabilities in multi-modal datasets.

The proposed guiding heuristic was compared against other methods, including the original \textsc{iForest} algorithm as well as several variations of it that introduce non-uniformly-random split choice heuristics, and was found to offer increased performance for the outliers of utmost interest (clustered outliers from multi-modal datasets), at the expense of degraded performance in other classes of outliers.

Compared to other methods, the proposed \textsc{FCF} algorithm not only showed increased performance as measured by ranking metrics, but also resulted in a more computationally-efficient procedure than other guiding heuristics, making it a promising algorithm for anomaly detection in datasets with multi-modal distributions.

\bibliographystyle{plain}
\bibliography{isomix}

\end{document}